\documentclass{article}

\usepackage{microtype}
\usepackage{graphicx}
\usepackage{subfigure}
\usepackage{stfloats}
\usepackage{booktabs} % for professional tables

\usepackage{hyperref}
\usepackage[comma,numbers,sort&compress]{natbib}

\usepackage{bm,amsbsy}  % allows for bold greek chars and symbols 
\usepackage{caption}
    \usepackage[preprint]{neurips_2019}

\usepackage{lipsum}
\usepackage[utf8]{inputenc} % allow utf-8 input
\usepackage[T1]{fontenc}    % use 8-bit T1 fonts
\usepackage{hyperref}       % hyperlinks
\usepackage{url}            % simple URL typesetting
\usepackage{amsfonts}       % blackboard math symbols
\usepackage{nicefrac}       % compact symbols for 1/2, etc.

\usepackage{amsmath}
\usepackage{wrapfig}
\usepackage{bbm}

\newcommand{\trp}{{^\top}} % transpose
\newcommand{\Nrm}{\mathcal{N}}

\newcommand{\vx}{\mathbf{x}}
\newcommand{\va}{\mathbf{a}}
\newcommand{\vb}{\mathbf{b}}
\newcommand{\vy}{\mathbf{y}}

\newcommand{\vmu}{\bm{\mu}}
\newcommand{\vlam}{\bm{\lambda}}
\newcommand{\vones}{\mathbf{1}}
\newcommand{\vX}{\mathbf{X}}
\newcommand{\bX}{\mathbf{X}}
\newcommand{\bW}{\mathbf{W}}
\newcommand{\vw}{\mathbf{w}}
\newcommand{\tbW}{\tilde{\mathbf{W}}}
\newcommand{\bY}{\mathbf{Y}}
\newcommand{\bK}{\mathbf{K}}
\newcommand{\bSigma}{\bm{\Sigma}}
\newcommand{\diag}{\operatorname{diag}}
\newcommand{\bH}{\mathbf{H}}

\newcommand{\vecc}{\textrm{vec}}

\newcommand{\LL}{\mathcal{L}}

\newcommand{\RR}{\mathbb{R}} %real numbers 

\usepackage{color,xcolor}

\title{Efficient non-conjugate Gaussian process factor models for spike count data using polynomial approximations}

\author{%
  Stephen L. Keeley \\
  Princeton Neuroscience Institute\\
  Princeton University\\
  Princeton, NJ 08542 \\
  \texttt{skeeley@princeton.edu} \\
  \And
  David M. Zoltowski \\
  Princeton Neuroscience Institute\\
  Princeton University\\
  Princeton, NJ 08542 \\
  \texttt{zoltowski@princeton.edu} \\
  \And
  Yiyi Yu \\
  Electrical and Computer Engineering\\
  University of California, Santa Barbara\\
  Santa Barbara, CA 93106 \\
  \texttt{yiyiy@ucsb.edu} \\
  \And
  Jacob L. Yates \\
  Center for Visual Science\\
  University of Rochester\\
  Rochester, NY 14627 \\
  \texttt{jyates7@ur.rochester.edu} \\
  \And
  Spencer L. Smith \\
  Electrical and Computer Engineering\\
  University of California, Santa Barbara\\
  Santa Barbara, CA 93106 \\
  \texttt{sls@ucsb.edu} \\
    \And
  Jonathan W. Pillow \\
  Princeton Neuroscience Institute\\
  Princeton University\\
  Princeton, NJ 08542 \\
  \texttt{pillow@princeton.edu} \\
}

\begin{document}

\maketitle

\begin{abstract}
%Factor analytic models provide a powerful statistical framework for identifying low-dimensional latent structure underlying high-dimensional data.
Gaussian Process Factor Analysis (GPFA) has been broadly applied to the problem of identifying smooth, low-dimensional temporal structure underlying large-scale neural recordings. However, spike trains are non-Gaussian, which motivates combining GPFA with discrete observation models for binned spike count data. The drawback to this approach is that GPFA priors are not conjugate to count model likelihoods, which makes inference challenging. Here we address this obstacle by introducing a fast, approximate inference method for non-conjugate GPFA models. Our approach uses orthogonal second-order polynomials to approximate the nonlinear terms in the non-conjugate log-likelihood, resulting in a method we refer to as \textit{polynomial approximate log-likelihood} (PAL) estimators. This approximation allows for accurate closed-form evaluation of marginal likelihoods and fast numerical optimization for parameters and hyperparameters. We derive PAL estimators for GPFA models with binomial, Poisson, and negative binomial observations and find the PAL estimation is highly accurate, and achieves faster convergence times compared to existing state-of-the-art inference methods. We also find that PAL hyperparameters can provide sensible initialization for black box variational inference (BBVI), which improves BBVI accuracy. We demonstrate that PAL estimators achieve fast and accurate extraction of latent structure from multi-neuron spike train data.
\end{abstract}
\newpage
\section{Introduction}

Recent advances in neural recording technologies have enabled the collection of increasingly high-dimensional neural data-sets. Making sense of such data requires new statistical methods for extracting shared latent structure underlying multi-neuron responses.  Factor models provide one popular approach to this problem \cite{byron2009gaussian,cunningham2014dimensionality,lakshmanan2015extracting,Archer15,wu2017gaussian}. These models seek to characterize the structure underlying neural data in terms of a small number of latent variables. These models have been widely successful in both uncovering interpretable structure from neural population data and providing insight into representations of stimulus input and behavior in population activity \cite{zhao2017variational, wu2017gaussian, zhao2019stimulus}. However, factor models can be cumbersome to learn when the prior distribution over the latent variables and the likelihood governing the observations are non-conjugate. This arises commonly for neural data, where binned spiking observations are best characterized by count models (e.g., binomial, Poisson, and negative-binomial).

Formally, latent factor models seek to explain shared structure underlying high-dimensional observations $(\mathbf{y_1,y_2,}\ldots,\mathbf{y_T}) \in \RR^{N\times T} $ in terms of low-dimensional latent variables $(\mathbf{x_1,x_2,}\ldots,\mathbf{x_T}) \in \RR^{P\times T}$, where $N>P$ and the observations are ordered sequentially in time from $t = 1$ to $t = T$.
%The goal of inference is to compute a posterior distribution over these latent variables given data, $p_\theta(\mathbf{x_t}|\mathbf{y_{1:T}})$. 
A popular approach is to model the time series of latent variables with a Gaussian process (GP), which makes few assumptions about latent trajectories beyond the fact that they evolve smoothly in time.  When combined with a Gaussian observations model, the resulting approach is known as Gaussian Process Factor Analysis (GPFA) \cite{byron2009gaussian}. Recent work  has extended GPFA to incorporate Poisson observations, which provides a more appropriate model for spike train data \cite{buesing2012spectral,macke2011empirical,zhao2017variational,wu2017gaussian, zhao2019stimulus}.
%has played a major role in neuroscience. This allows 
%with a linear dynamical system \cite{buesing2012spectral,archer2014low,kao2015single} or a Gaussian %Process  \cite{wu2017gaussian,byron2009gaussian,pfau2013robust}. The Gaussian Process (GP) approach 
%oreover, different GPFA implementations have different observation models: either Gaussian %cite{lawrence2004gaussian,byron2009gaussian} or Poisson .  
However, closed-form inference under GPFA models is only possible when the model likelihood and prior are conjugate. Consequently, Poisson and other non-conjugate models require approximations to fit hyperparameters or obtain parametric expressions for the posterior distribution over latents.

Here, we introduce a novel procedure for learning non-conjugate GPFA models with count observations, which we refer to as Polynomial Approximate Log-likelihood (PAL). This method exploits an idea for rapid inference in generalized linear models using so-called ``approximate sufficient statistics''  \cite{huggins2017pass,Zoltowski18nips}, and extends it to the latent variable model setting. The basic idea involves approximating the nonlinear terms in the model log-likelihood using orthogonal polynomials. When the polynomial approximation is second-order, the likelihood term can be explicitly marginalized to obtain a closed-form expression for the marginal likelihood, and an approximately Gaussian posterior distribution over the latents. We explicitly derive PAL estimators for three GPFA models with different count statistics. This includes the previously implemented Poisson count-observation GPFA model \cite{zhao2017variational,zhao2019stimulus}, as well as GPFA with binomial and negative-binomial observations. These three  distributions (binomial, Poisson, and negative-binomial) have different dispersion characteristics which reflect various spiking properties in neurons in different areas of the brain \cite{Charles18, goris2014partitioning, linderman2016bayesian}.

We compare our PAL approach to Black Box Variational Inference (BBVI), a state-of-the-art method for approximate inference in non-conjugate models that is renowned for its simplicity and adaptability \cite{ranganath2014black,Archer15,Gao15} and the variational latent Gaussian Process (vLGP) \cite{zhao2017variational}, a previous algorithm used for Poisson noise GPFA. We find that PAL estimation exhibits comparable performance to these methods, but PAL compares favorably to both of them in that it provides a closed-form expression for marginal likelihood that can be optimized directly; it therefore requires no careful tuning of learning rates, number of Monte Carlo samples, or stopping criteria, and does not suffer from high-variance estimates due to sampling-based evaluation of marginal likelihood. We also find that, in each case, PAL is faster than these existing algorithms and can accurately recover latent structure in simulated neural data. 

 We further demonstrate that PAL hyperparameters can be used to initialize BBVI to stabilize and improve inference. We use this combined BBVI + PAL on two different multi-neuron datasets, one from mouse visual cortex and one from primate parietal cortex, under three different choices of count model (binomial, Poisson, and negative binomial). We show that PAL initialized BBVI performs as good or better than BBVI alone. The PAL approach therefore offers a promising avenue for future work on non-conjugate models that arise frequently in the analysis of biological and other data.

%We show that that count-GPFA models generally outperform standard Gaussian GPFA for extracting latent structure from spike train data.
 %Overall, the PAL method provides two contributions to non-conjugate GPFA inference: (1) it is an accurate closed form method of marginalizing count-GPFA models and (2) in situations non-linearities cannot be well-approximated with quadratic functions, PAL provides a sensible initialization to manage BBVI. 
%GPFA count models are widely applicable to high-dimensional time-series data with count observations. Likewise, 

%and overcome the need for closed form expressions of the posterior. They achieve this by approximating the posterior distribution $p_{\theta}(\mathbf{x}|\mathbf{y})$% = p_{\theta}(\mathbf{x},\mathbf{y})/p_{\theta}(\mathbf{y})$  
%with a well-behaved variational distribution $q_{\phi}(\mathbf{x}|\mathbf{y})$. But even with this approximation, closed form variational inference requires calculating the expectation under $q_{\phi}(\mathbf{x}|\mathbf{y})$ of the joint distribution $p_{\theta}(\mathbf{x},\mathbf{y})$, which may be intractable. As such, 'black-box' variational inference methods (BBVI), which work via sampling of the joint distribution, avoid the need for the calculation of $p_{\theta}(\mathbf{x},\mathbf{y})$, and promote a flexible inference procedure \cite{ranganath2014black}.

\begin{figure*}[t] 
   \centering
   \includegraphics[width=1\textwidth]{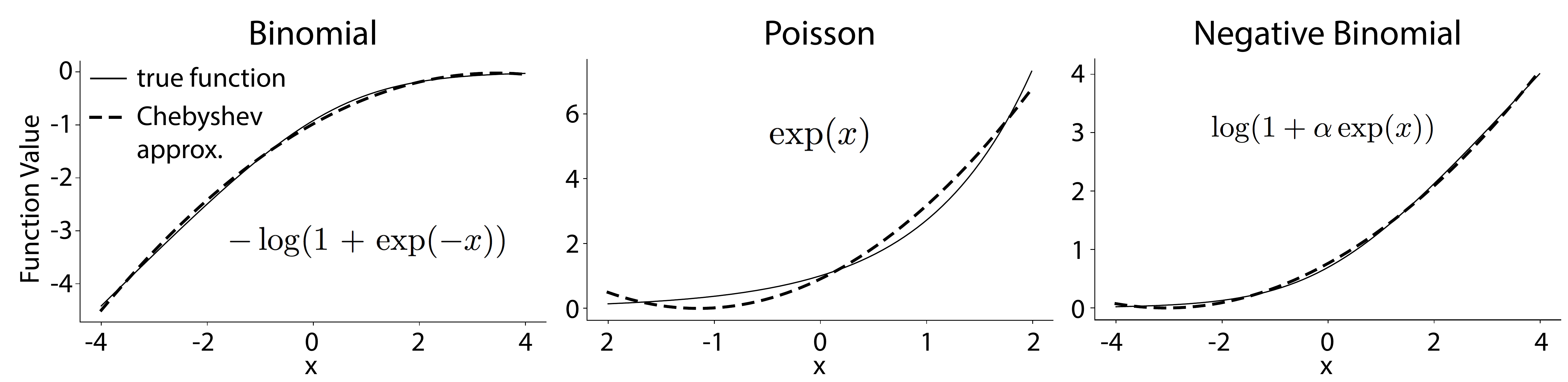}   
   \caption{Comparison of nonlinear term found in the log-likelihood for binomial, Poisson, and negative-Binomial observation models (solid) with corresponding second-order Chebyshev approximation (dashed).}
   \label{fig:cheby}
\end{figure*} 

%\vspace*{-40 px}

\section{Count-GPFA models}

Consider a dataset consisting of count observations from $N$ neurons over $T$ time bins, $\bY \in \mathbb{N}^{N\times T}$.  The count-GPFA model seeks to describe these data in terms of a nonlinearly transformed linear projection of  lower-dimensional latent variable $\bX \in{\RR}^{P \times T}$, $P<N$, where each latent variable evolves according to an independent Gaussian process.  Thus the timecourse of the $j$'th latent variable, which forms the $j$'th row of $\vX$, has a multivariate normal distribution:
\begin{equation}
  \label{eq:1}
\vx_j \sim \Nrm(0,K_j),
\end{equation}
where each $K$ is a $T \times T$ covariance matrix whose $(t,t')$'th  entry is given by the covariance function $k(t,t')$. In this paper, we use the common Gaussian or ``squared exponential'' covariance function:  $k(t,t') =  \exp(-(t-t')^2/(2\ell^2))$, which is governed by a single hyperparameter, the ``length scale'' $\ell$, which controls smoothness of the latent process. 

The count-GPFA observation model can then be written:
\begin{equation}\label{eq:obs}
  \bY|\bW, \bX \sim\ \mathcal{P}(f(\bW \bX))
\end{equation} 
where $\bW \in \RR^{N\times P}$ is a loading matrix, $f(\cdot)$ denotes a nonlinear function that transforms $\bW \bX$ to the appropriate range for a count random variable (e.g., the non-negative reals), and $\mathcal{P}$ denotes a probability distribution for count data.

Fitting the count-GPFA model to data involves inferring the loading weights $\bW$ and hyperparameters $\theta = \{\ell_1, \dots \ell_j \}$ via numerical optimization of the marginal likelihood:
\begin{equation} \label{eq:margli}
    P(\bY | \bW,\theta) = \int P(\bY | \bW, \vX) P(\vX|\theta) d\vX.
\end{equation}
However, non-conjugacy of the count model likelihood $P(\bY|\bW,\vX)$ and Gaussian prior over latents $P(\vX|\theta)$ means that this integral cannot be computed in closed form. Likewise, the posterior distribution over latents given the data, given by:
%\begin{equation}
$P(\vX | \bY,\bW,\theta) = P(\bY|\vX,\bW) P(\vX|\theta) / P(\bY|\bW,\theta)$,
%\end{equation}
has no closed form expression, where the desired normalizing constant is the marginal likelihood. Fitting and inference therefore rely on approximate inference methods.

\section{Polynomial Approximate Log-likelihood (PAL)}

Here we propose Polynomial Approximate Log-likelihood (PAL), an approximation scheme for efficient inference in non-conjugate Gaussian latent variable models.
%Gaussian settings that we call the Polynomial-Approximate Log-likelihood (PAL) method.
The core idea is to approximate terms in the observation model log-likelihood that are nonlinear in $\vX$ using orthogonal polynomials.  Our approach is inspired by recent work on ``polynomial approximate sufficient statistics'' for generalized linear models (PASS-GLMs) \cite{huggins2017pass,Zoltowski18nips}. In that work, the $\vX$ were observed regressors, and the method provided so-called ``approximate sufficient statistics'' that could be computed with a single pass over the data.

Here, the $\vX$ are (unobserved) latent variables instead of regressors, and the goal of the approximation is efficient marginalization rather than a set of sufficient statistics.
We consider second-order polynomial approximations to the log-likelihood, which allow for analytic marginalization over latents.  PAL therefore enables closed-form evaluation of the approximate marginal likelihood, allowing efficient optimization of parameters and hyperparameters.
 
We derive PAL estimators for GPFA under three different non-conjugate observation models: binomial, Poisson, and negative binomial (NB). These models range from under-dispersed or ``sub-Poisson'' for binomial to overdispersed or ``supra-Poisson'' for NB,  thus spanning the range of dispersion behaviors found in different brain areas \cite{Kara00,Maimon09,Pillow12,goris2014partitioning,Gao15,Charles18}.  
%Each model involves different nonlinear log-likelihood terms, which we derive below for each model. 

All PAL count-GPFA models have the same general form for the approximate log marginal likelihood (log evidence):
\begin{equation}
\mathcal{E}(\vy|\bW,\theta) \approx \frac{1}{2} \log |\bSigma| + \frac{1}{2} \vmu^\top \bSigma^{-1} \vmu - \frac{1}{2} \log |\bK|,
\label{eqn:approx}
\end{equation}
where $\bSigma$ denotes an approximate posterior covariance and $\vmu$ denotes an approximate posterior mean, and $\bK$ is the prior covariance over all latents (a block-diagonal matrix, with one block for each latent). The form of the first two terms varies across models, which we derive for three specific models below. See Table~\ref{pal-table} for a summary of the results for all count-GPFA models. For clarity, we define $\bH = \bSigma^{-1} - \bK^{-1}$ in this table to succinctly present approximate posterior covariances.

\begin{table*}[!ht]
\begin{center}
%\begin{small}
%\begin{sc}
\begin{tabular}{rccccc}
\toprule
& & binomial  & Poisson & negative binomial & \\
\midrule
spike rate $\lambda_{it}$& &  $n \sigma(\vw_i\trp \vx_t)$ & $\exp(\vw_i\trp \vx_t)$ & $\exp(\vw_i\trp \vx_t)$  \\
 nonlinear term & & $- \log(1+e^{-x})$ & $e^{x}$& $ \log(1+\alpha e^{x})$   \\
$\bH$  & & $2n\tbW^\top \diag(\va) \tbW $& $2  \tbW^\top \diag(\va) \tbW $ & $2 \tbW^\top  \diag( (\alpha^{-1} + \vy) \circ \va)  \tbW$\\ 
posterior mean $\vmu$ & &$ \bSigma \tbW^\top  (\vy -n - n \vb)$& $\bSigma \tbW^\top  (\vy - \vb)$& $\bSigma \tbW^\top  (\vy - \vy \circ \vb - \alpha^{-1} \vb)$ \\
\bottomrule
\end{tabular}
%\end{sc}
%\end{small}
\end{center}
\caption{Summary of PAL expressions for count-GPFA models. Top line gives the spike rate of neuron $i$ at time $t$ given the latent vector $\vx_t$ and loading weights $\vw_i$ for neuron $i$. Second line gives the nonlinear term of the log-likelihood that must be approximated under PAL.  The third row, $\bH$ is defined by $\bH = \bSigma^{-1} - \bK^{-1}$, which succinctly presents posterior covariances, and the fourth line $\vmu$ shows approximate posterior means.}
\label{pal-table}
\end{table*}

\subsection{PAL for Poisson-GPFA}

We begin with the Poisson observation model, which is the most common model for spike counts and a popular choice for latent variable models of spike train data \cite{wu2017gaussian,zhao2017variational,duncker2018temporal}. For this model, spike count $y$ given a spike rate parameter $\lambda$ is distributed according to:
\begin{equation}
    P(y|\lambda) = \tfrac{1}{y!} (\Delta \lambda)^{y} e^{- (\Delta\lambda)},
\end{equation}
where $\Delta$ is the time bin size (which we set here to 1, resulting in spike rates in units of spikes/bin).  We use an exponential nonlinearity from latents to spike rates, so the vector of spike rates at time $t$ is:
\begin{equation}
    \vlam_t = \exp(\bW\vx_t).
\end{equation}
This choice of nonlinearity gives rise to a log-likelihood with a single nonlinear term, although other nonlinearities can be considered \cite{Zoltowski18nips}. 

The Poisson log-likelihood for the entire dataset can be written conveniently in vector form as:
\begin{equation}
  \LL(\vy,\vx|\tbW) =  \vy^\top \tbW \vx - \mathbf{1}^\top \exp(\tbW \vx) + const
  %-\frac{1}{2} \vx^\top \bK^{-1} \vx - \frac{1}{2} \log | \bK |
\end{equation}
where $\vy=\vecc(\bY)$ is a $NT \times 1$ vector of concatenated spike count observations from all $N$ neurons and $T$ time bins, $\vx=\vecc(\vX)$ is a $PT \times 1$ vector of concatenated latent vectors across $P$ latent time series, $\tbW = \bW\otimes\mathbf{I}_T$ is a $NT \times PT $ Kronecker-structured matrix, and $\mathbf{1}$ is a length-$NT$ vector of ones.  

%Here, $\bK$ is the block-diagonal covariance of the concatenated $\vx$, and  we have ignored a constant that involves the terms that do not depend on the $\tilde \bW \vx$. 

The only nonlinear term in the log-likelihood is the exponential term $\exp(\tilde \bW\vx)$. We therefore approximate the exponential function with a second-order polynomial:
\begin{equation}
\exp(x) \approx a x^2 + b x + c,
\end{equation}
with coefficients $a$, $b$, and $c$ given by a Chebyshev polynomial approximation to $\exp(x)$ over an interval $\psi = [x_0,x_1]$, which we set independently for each neuron \cite{mason2002chebyshev,Zoltowski18nips}. We use Chebyshev polynomials because they provide efficient near-minimax polynomial approximations \cite{huggins2017pass}. Specifically, we computed the truncated Chebyshev expansion of the exponential $\exp(x) = \sum_{m=0}^{2} = \beta_m T_m$ where $T_m$ is the degree-$m$ Chebyshev polynomial of the first kind over $[x_0,x_1]$ and $\beta_m$ are the expansion coefficients over that interval. The coefficients $a$, $b$, and $c$ are given by collecting the terms to rewrite the expansion in the monomial basis. 

We selected the interval $[x_0,x_1]$ independently for each neuron by computing the log of the mean firing rate of each neuron, $\log \lambda_{i}$. Since the nonlinearity is over the input $\bW \vx$, and the firing rate is $\lambda = \exp(\bW \vx)$, we take the log of $\lambda_i$ as we wish to center the nonlinear approximation at the center of the empirical neuronal rate to maximize accuracy. See Figure \ref{fig:cheby} as an example of a range centered at 0, corresponding to a simulated GP drawn with mean 0.  We then chose the limits of the range to be $[\log \lambda_i - 2, \log \lambda_i + 2]$, resulting in an approximation range extending from $e^{-2}$ to $e^2$ times the mean firing rate. We found that this range balanced coverage in firing rate space with approximation accuracy. After selecting the range centers for each neuron, we computed the polynomial coefficients $( a_i, b_i, c_i)$ for neuron $i$ by gridding the interval of interest at a resolution of $dx = 0.01$ and solving for the coefficients that minimize the least squares between the true function and its polynomial approximation. For more detail, see \cite{Zoltowski18nips}.

Given coefficients for each neuron, the exponential term in the Poisson log-likelihood can be approximated:
%\begin{equation}
%\begin{aligned}
%    &\vones\trp \exp(\tbW \vx)\\ &\approx \sum_{t=1}^T \sum_{i=1}^N \Big( a_i (\bW \vx_t)_i \circ (\bW \vx_t)_i  + b_i (\bW \vx_t)_i + c_i\Big) \\
%    &= \vx\trp {\tbW}^\top \diag(\va) \tbW \vx  + \vb^\top \tbW +  const,
%\end{aligned}
%\end{equation}
\begin{align}
    & \vones\trp \exp(\tbW \vx) \nonumber \\  & \qquad \approx \sum_{t=1}^T \sum_{i=1}^N \Big( a_i (\bW \vx_t)_i \circ (\bW \vx_t)_i  + b_i (\bW \vx_t)_i + c_i\Big) \nonumber \\
     & \qquad = \vx\trp {\tbW}^\top \diag(\va) \tbW \vx  + \vb^\top \tbW +  const,
\end{align}
where $\circ$ denotes Hadamard (element-wise) multiplication, and the second line involves the concatentation of the polynomial coefficients for each neuron and time bin: $\va = [a_1 \vones, \ldots a_N \vones]\trp$,  $\vb = [b_1 \vones, \ldots b_N \vones]\trp$, and we can ignore the constants $c_i$.

%1)} \mathbf{1},...,a_i^{(N)} \mathbf{1}]^\top$ is the concatenation of the approximation coefficients for each time bin and neuron. 

We now substitute the polynomial approximation into the log-likelihood and add the log prior, giving:
%\begin{equation}
\begin{align}
  & \LL(\vy,\vx|\tbW,\theta) \approx \nonumber \\ 
  & \qquad \qquad \vy^\top \tbW \vx - \vx^\top \tbW^\top \diag(\va) \tbW \vx  \nonumber \\ & \qquad \qquad - \vb^\top \tbW \vx -\frac{1}{2} \vx^\top \bK^{-1} \vx - \frac{1}{2} \log | \bK |.
\end{align}
Since this approximation is quadratic in $\vx $ we can exponentiate and then analytically marginalize $\vx$ to obtain an approximation to the log-likelihood that follows equation (\ref{eqn:approx}) where:
\begin{align}
    \bSigma^{-1} &= 2  \tbW^\top \diag(\va) \tbW+ \bK^{-1} \\
    \vmu &= \bSigma \tbW^\top (\vy - \vb),
\end{align} 
and we have dropped terms that do not depend on $\tbW$ or $\theta$.

\begin{figure*}[t] %  figure placement: here, top, bottom, or page  
   \centering
   \includegraphics[width=1\textwidth]{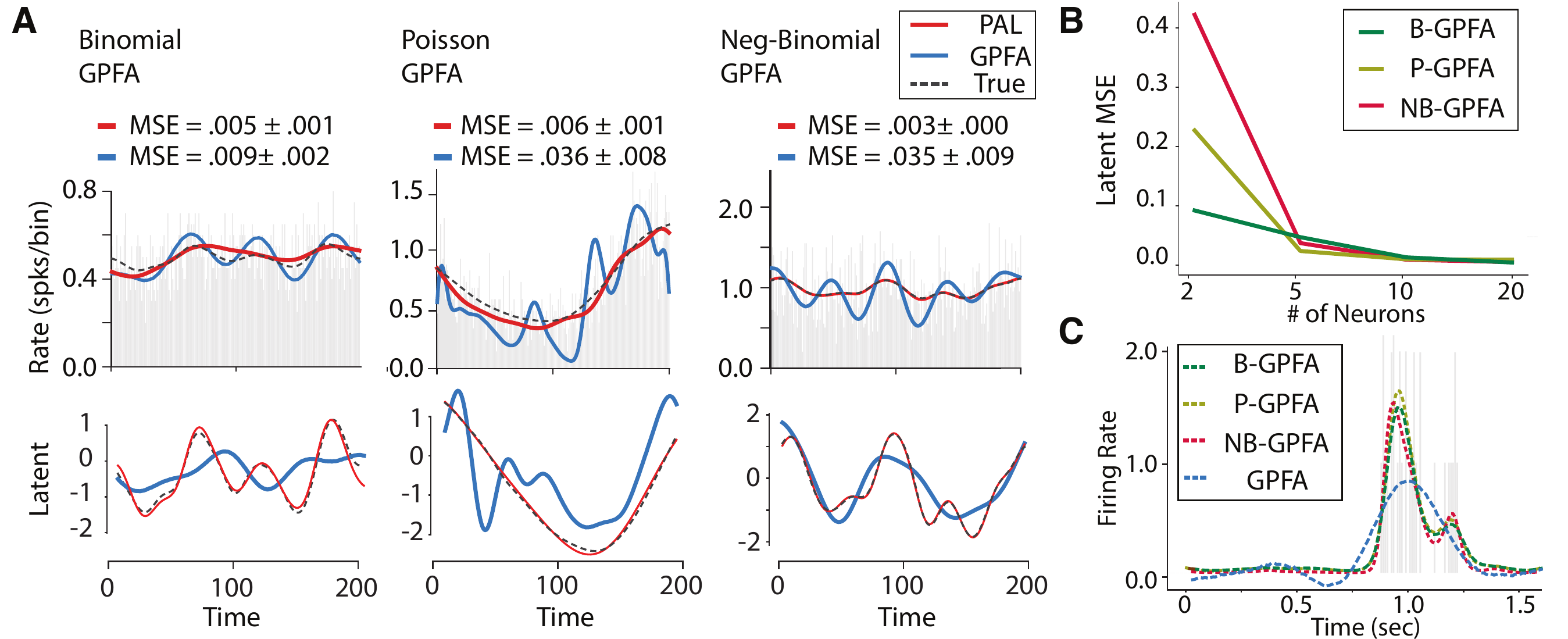}       \vskip -0.1in
      \caption{PAL inference for population data from binomial, Poisson, and negative binomial GPFA models. \textbf{A.} One example simulated neuron (out of 20) are shown for each model.  Inferred rate for neurons for PAL inference compared to GPFA (Top). Latent trajectories recovered for each model (bottom). \textbf{B.} Error of recovered latent structure falls to zero with increasing numbers of observed neurons, as expected. \textbf{C.} Example PAL fits for all count-GPFA models compared to standard GPFA for spiking data from an example neuron. Light grey histogram denotes spike-count observations.}
   \label{fig:PAL_ex}
\end{figure*}

\subsection{PAL for Binomial-GPFA} Deriving the PAL estimator for a binomial observation model follows a similar logic to the Poisson case. Recall that for binomial model, spike count $y$ is distributed according to
: \begin{equation}
  P(y|p, n) =  \binom{n}{y} p^y(1-p)^{(n-y)}.
\end{equation}
%where $p$ is the ``probability of success'' and $n$ is the ``number of trials'' parameter.
For this model, we map latents through a sigmoidal nonlinearity, $\sigma(x) = 1/(1+\exp(-x))$, to obtain the binomial parameter $p$, and we set the number-of-trials parameter, $n$, to be the maximum number of observed spikes in a single time bin. The vector of spike rates at time $t$ for this model is thus given by:
\begin{equation}
  \label{eq:2}
  \vlam_t = n \sigma(\bW \vx_t).
\end{equation}

We can write the log-likelihood in vectorized form as:
\begin{equation}
\begin{aligned}
  \LL(\vy|\vx, \tbW)
  % &=  \vy^\top \log(\sigma(\tbW\vx)) + (\tvn-\vy)^\top \log(1-\sigma(\tbW\vx)) + const \\
  &= (-n+\vy) \tbW\vx - \\ &n \log(1+\exp(-\tbW\vx)) + const
    \end{aligned}
\end{equation}
where we have ignored terms that do not depend on $\tbW \vx$.

The problematic term here is the nonlinear second term, $\log(1+\exp(-x))$, which we approximate, as before, using a second-order Chebyshev polynomial approximation. In this case, we choose the center of the non-linearity to be the inverse sigmoid function of the empirical mean rate for each neuron $\sigma^{-1}(\lambda_i)$. We use a range of $[\sigma^{-1}(\lambda_i)-4,\sigma^{-1}(\lambda_i)+4]$ for the Chebyshev approximation. As in the Poisson case, we do this so the range for each neuron is centered at the average empirical value of input term to the non-linearity, $\tbW \vx$. %consistent with previous work in the logistic regression case \cite{huggins2017pass}. 
The resulting approximation to the log-likelihood is:
\begin{equation}
\begin{aligned}
  \label{eq:4}
    \LL(\vy|\vx,\tbW)  &\approx  -n\vx\trp {\tbW}^\top \diag(\va) \tbW \vx  \\ &+(\vy - n -n\vb)^\top \tbW \vx +  const
    \end{aligned}
\end{equation}

As in the Poisson case, we can add the log-prior to the above expression, exponentiate and marginalize over $\vx$ to obtain an approximation to the log marginal likelihood in the same form as equation~(\ref{eqn:approx}). In this case, we obtain matrix and vector terms:
\begin{equation}
\begin{aligned}
\bSigma^{-1} &= 2n  \tbW^\top \diag(\va) \tbW + \bK^{-1}\\
\vmu &= \bSigma \tbW^\top (\vy -n - n \vb).
    \end{aligned}
\end{equation}

\subsection{PAL for negative-binomial GPFA}
Lastly, we consider a negative binomial observation model, which covers the over-dispersed spike responses \cite{Pillow12,linderman2016bayesian,goris2014partitioning}. For negative-binomial GPFA, we parametrize the negative binomial distribution in terms of mean parameter $m$, and scale parameter $r = 1/\alpha$:
\begin{equation}
P(y|m, \alpha) = \frac{\Gamma(y+\alpha^{-1})}{\Gamma(\alpha^{-1})\Gamma(y+1)}\big(\frac{1}{1+\alpha m}\big)^{\alpha^{-1}}\big(\frac{\alpha m}{1+\alpha m}\big)^{y}
\end{equation}
This form of the distribution maps to the standard negative-binomial distribution, $p(y|p,r) = \binom{y+r-1}{y} (1-p)^r p^y$, via $p = \frac{r}{m+r}$.
\begin{figure*}[t] %  figure placement: here, top, bottom, or page  
   \centering
   \includegraphics[width=1\textwidth]{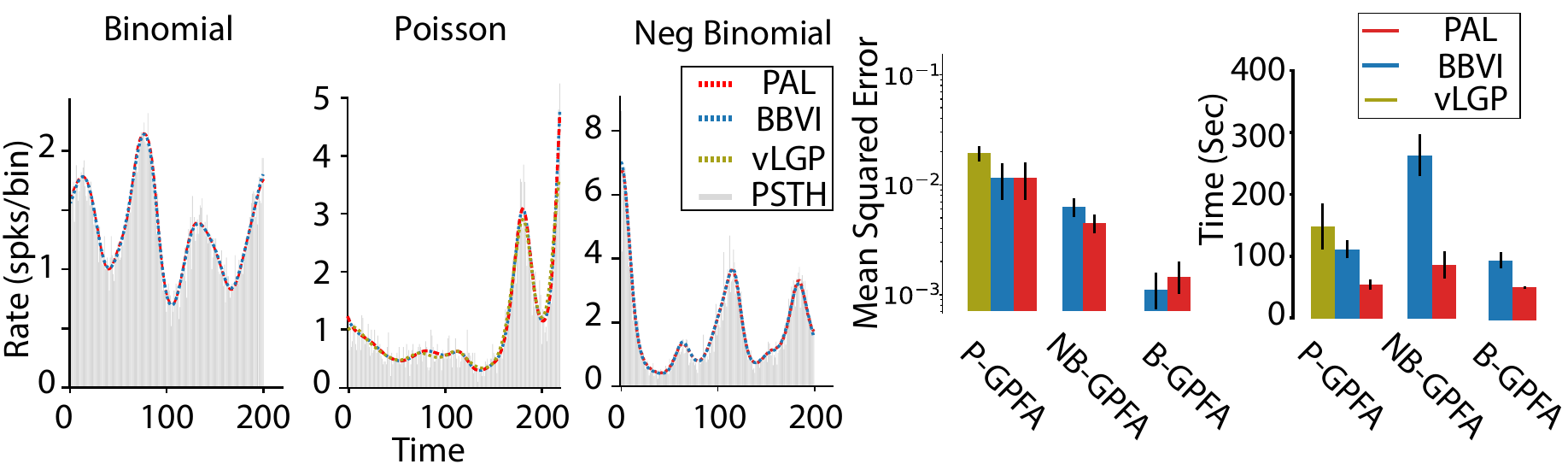}   
      \caption{Comparison of vLGP, BBVI and PAL-based estimation in count-GPFA models. (\textbf{Left}) Reconstructed spike rates for each inference procedure, plotted alongside the spiking activity summed across trials (PSTH). Here, PAL-learned hyperparameters followed by a MAP estimation (red), BBVI (blue) and vLGP (yellow) all yield very similar and highly accurate spike rate reconstructions. Each of these closely match true spiking data (gray PSTH). (\textbf{Middle}) MSE of reconstructed rates shows good performance for all methods. (\textbf{Right}) PAL is faster than competing algorithms.}
   \label{fig:P-GPFA_BBVI_ex}
\end{figure*}
Parameterizing the negative binomial model this way makes for a simple expression of the expected spike count, which is equal to the model parameter $m$. Let us define this mean rate in the factor analytic framework as $m = \exp(\tbW\vx)$. This allows us to write the log-likelihood in vector form as:
\begin{equation}
\begin{aligned}
\LL(\vy|\tbW,&\vx, \alpha) =  \vy^\top\tbW \vx - \\ &(\alpha^{-1}+\vy^\top)\log(1+\alpha \exp(\tbW \vx)) + const.
    \end{aligned}
\end{equation}
To derive a PAL estimator, we use a quadratic approximation to the nonlinear term $\log(1+\alpha\exp(x))$ on a per-neuron basis. We set $\alpha = 1$ for simulations, but this quantity may be learned in an outer loop. We choose the center of the nonlinear range to be the same as in the Poisson case, with the center value being the log of the mean firing rate of the neuron (see right panel of Figure \ref{fig:cheby} for example of centering with an average log-rate of 0). The range limits are $[\log \lambda_i - 4, \log \lambda_i + 4]$, where $\lambda_i$ is the average value of $m$ across time, per neuron. Here, a wider range can be used as this nonlinearity is accurately captured by the quadratic approximation. As in the previous cases, we obtain a quadratic approximate log-likelihood which has the following form:
\begin{equation}
\begin{aligned}
  \label{eq:logliNB}
    \LL(\vy|\vx,\tbW, \alpha)  \approx  &-\vx\tbW^\top  \diag( (\alpha^{-1} + \vy) \circ \va)  \tbW\vx \\
    &+ (\vy - \vy \circ \vb - \alpha^{-1} \vb)^\top \tbW \vx +  const
\end{aligned}
\end{equation}
We then add the log prior and marginalize $\vx$ to obtain an approximation to the log marginal likelihood for negative-binomial GPFA that follows the same form as equation~(\ref{eqn:approx}) with 
\begin{align}
\bSigma^{-1} &= 2 \tbW^\top  \diag((\alpha^{-1} + \vy) \circ \va)  \tbW + \bK^{-1}\\
\mathbf{\mu} &= \bSigma \tbW^\top (\vy - \vy \circ \vb - \alpha^{-1} \vb)
\end{align}

A summary of the features of all count GPFA models is given in Table 1. This table lists the nonlinear term for each model, the expected number of spikes for the $i$th neuron as a function of the latents, $\bX$, loadings matrix $\bW$, and the mean and covariance of the polynomial-approximated marginal distribution. We use $n_i$ to refer to the maximal spike count for neuron $i$, and $\mathbf{w}_i$ to denote the $i$th column of $\bW$.

\subsection{Evaluating PAL performance}

To assess the accuracy of the PAL estimator, we first analyzed its performance on simulated data. For 20 trials with 200 time points, we simulated count observations from 20 neurons with 2 latent processes with length scales $\ell_1 = 15$ and $\ell_2 =60$ and each entry of $\bW$ drawn uniformly in $[0,2]$. %The PAL method provides accurate latent variable recovery as demonstrated by the reconstructions shown at the bottom of Figure \ref{fig:PAL_ex}. 
We then fit each model by directly optimizing equation~\ref{eqn:approx} to obtain parameter estimates $\hat{\bW}$ and hyperparameter estimates $\hat{\ell}$. Conditioned on these estimates, we then maximized the conditional posterior to obtain $\hat{\mathbf{X}}_{MAP}$,  the MAP estimate of the latent process. As a control, we compared PAL performance to standard Gaussian-noise GPFA. 

We found that the rates estimated using this procedure were similar to the true model rates and showed substantial improvement above Gaussian GPFA (Figure \ref{fig:PAL_ex}A, top). Additionally, PAL inference accurately captures latent structure (Figure \ref{fig:PAL_ex}A, bottom), whereas GPFA cannot. To identify latent structure in these simulated data, we regress learned latents onto the true latents as latent factors models are identifiable only up to a rotation matrix. Accurate identification of latent structure is a primary feature of this inference procedure, as latents have functional importance in neuroscience settings \cite{byron2009gaussian, zhao2017variational, duncker2018temporal}. %Moreover, because traditional GPFA has no nonlinearities present, the recovered latent structure for traditional GPFA is always very distinct from true count-GPFA latent structure, even if reconstructed neural rates are reasonably close to the true rates. 

We additionally demonstrate PAL's accuracy by showing error of recovered latent structure as a function of the number of observed neurons. For each count-GPFA model, as we consider more and more data (from 2 to 20 neurons), PAL more and more accurately recovers latent structure, as expected (Figure \ref{fig:PAL_ex}B). Fits for real neural data are shown for an example neuron in Figure \ref{fig:PAL_ex}C. This is the first half of a trail for an example neuron from the rodent dataset (for more information, see section 5). PAL fits to count-GPFA better describe the neural spike-count data than standard GPFA.  The background histogram in light grey in Figure \ref{fig:PAL_ex}C shows the true spike counts, and each of the dotted lines show the estimated neural firing rates under each model. Standard GPFA inference problematically yields negative rates and fails to capture the quick changes in firing rate. 
\begin{figure*}[t] 
  \centering
  \includegraphics[width=1\textwidth]{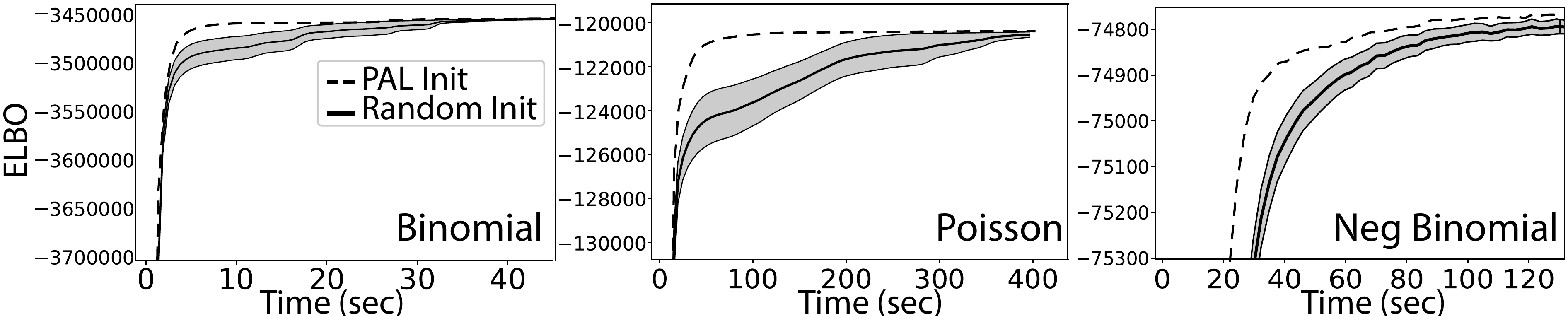}   
  \caption{Optimization time for full BBVI from either a random initialization or PAL initialization for all count-GPFA models. Data is shown as mean and standard error for 10 trajectories. 
  }
  \label{fig:bbvi_elbos_all}
\end{figure*}

\section{Comparison to other approaches}

Variational inference \cite{blei2003latent} represents a common alternate approach to performing inference in non-conjugate factor models. This approach has been previously used in the setting of Poisson-GPFA \cite{zhao2017variational, duncker2018temporal}, and could in principal be used for the two other count-GPFA models we have introduced here.  We therefore show comparisons of PAL to a variant of variational inference called black-box variational inference (BBVI), which uses Monte-Carlo samples to approximate the expectation term in the ELBO \cite{kingma2015variational}. We additionally compare to an existing inference method available for Poisson GPFA called the variational Latent Gaussian Process (vLGP) \cite{zhao2017variational}.
% \begin{multline}\label{eq:elbo}
%  \mathbb{E}_q[\log(p(\mathbf{y}|\mX,\mathbf{W}))] \approx \\  \frac{N}{M}\sum_{i=1}{M}\log  p(\mathbf{y}|\mathbf{W}, \mX = f(\epsilon, \phi))
% \end{multline} 
% We optimize the model and variational parameters using samples from $q_\phi(X)$. To reduce variance, we parameterized $q_\phi(X)$ as a differentiable function of standard normal random variables $f(\epsilon, \phi)$ and computed gradients with respect to $\phi$ using the local reparameterization trick \cite{kingma2015variational}. 
%The "black box" approach refers to the fact that $\mathbb{E}_{q_\phi}[\log(p_{\theta}(y|x,w))]$ is not calculated directly, but rather is approximated using Monte-Carlo samples. We parameterize $q_\phi$ to be a differentiable function of standard normal noise $f(\epsilon, \phi)$ such that
%\begin{equation}\label{eq:elbo2}
% \mathbb{E}_{q_\phi}[\log(p_{\theta}(\mathbf{y}|\bX,\mathbf{W}))] \approx \\  \frac{N}{M}\sum_{i=1}{M}\log  p_{\theta}(\mathbf{y}|\mathbf{W}, \bX = f(\epsilon_i, \phi)).
%\end{equation} 
%Because the learning of variational parameters $\mathbf{x}$  is typically subject to instability via the Monte-Carlo estimation step, we employ the local reparameterization trick, which greatly reduces the variance of the estimates by sampling over a smaller number of dimensions \cite{kingma2015variational}. 

%Through optimization of the ELBO (\ref{eqn:elbo}) we identify both the latents $\bX$ and hyperparameters $\mathbf{W, \ell}$, via optimization of the variational distribution $q_\phi$. 

On simulated data, BBVI, PAL and vLGP inference procedures achieve highly accurate reconstructions of true spike rates. The rate reconstructions of three example neurons for each model are shown on the left panel of Figure \ref{fig:P-GPFA_BBVI_ex}. Here, each model nearly perfectly predicts the simulated neural data. Average MSE across all neurons for these count GPFA models are shown in the middle panel of Figure \ref{fig:P-GPFA_BBVI_ex}. Though all inference methods achieve accurate results, times-to-convergence are faster and much more stable using the PAL approach (Figure \ref{fig:P-GPFA_BBVI_ex}, right panel). The time to converge is determined by the average times-to-convergence of ten runs of each optimization procedure. In the BBVI case, convergence was determined when the ELBO was within 99.8\% of the maximal ELBO value identified. For occasional BBVI runs for each count model, this value was not achieved for the duration of the inference procedure, as the the ELBO was stuck at a local maxima. These convergence times were discarded when calculating the mean convergence time, and demonstrative of the irregularity of the BBVI inference procedure. For vLGP, we set the number of maximum iterations to 50, and the minimum to 10. The algorithm typically did not converge before the maximum number of iterations was up. We also note a slight accuracy improvement of PAL compared to vLGP. However, it is important to note that vLGP assigns a per-trial latent whereas our algorithm assumes the latent is same across all trials. This is possibly a confounding factor when comparing performance time of vLGP and PAL, both of which achieve high accuracy. 

% Controlling the variance of stochastic gradients of the ELBO in BBVI is an active area of research. The use of sampling in the optimization poses considerable challenges in convergence, and as such a variety of techniques have been introduced to reduce variance in the gradient estimates, including use of a natural gradient for optimization \cite{hoffman2013stochastic,salimbeni2018natural}, reformulating the gradient estimator \cite{roeder2017sticking}, and the reparameterization trick \cite{kingma2015variational, rezende2014stochastic}. Despite employing the reparameterization trick, we have likewise found for our GP latent models that BBVI is similarly unstable and can take widely variable amounts of time to converge depending on initialization of the parameters and the stochastic trajectory (Figure \ref{fig:P-GPFA_BBVI_ex}). Further, we use Adam optimization \cite{kingma2014adam}, which will occasionally find local maxima and retain a low value of the ELBO estimate for a substantial period of time. Despite these shortcomings, we have found that if the sampling-based inference is allowed to run for long enough, BBVI recovers true latent structure. 

\subsection{PAL initialization for BBVI}
The PAL method can additionally be used in conjunction with BBVI to speed up and improve inference. Because the PAL method involves approximating a nonlinearity in a specified range with a quadratic, if the input ($\bW\vx$) is not within the range of the Chebyshev approximation the estimate will be inaccurate.  Moreover, there is a significant limitation using BBVI. In particular, the use of sampling in the optimization poses considerable challenges in convergence. In fact, this is a well-known problem, and a variety of techniques have been introduced to reduce variance in the gradient estimates \cite{roeder2017sticking,hoffman2013stochastic,salimbeni2018natural}. We offer an alternative solution; we can overcome the limitations of both BBVI and PAL by combining them.

Initializing the BBVI algorithm with the hyperparameters provided by PAL optimization of equation~\ref{eqn:approx} allows for a rapid and stable BBVI. That is, instead of following up our PAL hyper-parameter identification with a MAP estimation of the latents, we use it to seed an approximate accurate hyperparameters to BBVI. This procedure is more stable than full BBVI with random initial hyperparameters. We demonstrate this for all count-GPFA models. Figure \ref{fig:bbvi_elbos_all} shows the evolution of the ELBO in time during optimization for all models on simulated data (though the same effect is observed in real data). In each case, BBVI is run 10 times, either initializing randomly or initializing at the PAL-optimal hyperparameters ($\ell$ and $\bW$). Standard error is shown in grey for the random initialization, but not shown for PAL-initialized optimization, as this trajectory follows nearly identically for each run. An initial sharp increase in the ELBO is always observed in all models, as here latent structure is approximately identified, but hyperparameters are tuned at the end of the BBVI optimization procedure. Here, we have cut off the initial rise in ELBO for clarity. Figure \ref{fig:bbvi_elbos_all} demonstrates the end of the optimization procedure, where randomly initialized BBVI attempts to find hyperparameters along varying trajectories, but PAL initialized BBVI quickly converges to high ELBO values. %This is observed despite using state-of-the-art methods for BBVI inference \cite{kingma2014adam,kingma2015variational}. %In fact, for many of the trajectories tested, the optimum ELBO value was not within 99.9\% of the maximum value seen using PAL-initialization for the entire duration of the time ran. 

%The BBVI inference approach is uncertain, and it can be impossible to tell when the procedure has achieved a true maximum, or is stuck in a local optimum. PAL-initialization overcomes this limitation, and our polynomial approximation provides a principled method to initialize BBVI at good hyperparameter estimates. 
Thus, initializing BBVI with PAL hyperparameter estimates avoids local optima. This PAL initialization procedure can therefore be considered alongside other methods for providing a way to stabilize and improve BBVI. 

\section{Applications to neural data}

 To examine the performance of these methods on real data, we applied the binomial, Poisson, and negative binomial count-GPFA models to neural data sets from two different species. We compared the three approaches outlined: PAL followed by a MAP estimate (section 3), BBVI, and  PAL-initialized BBVI (section 4). We tested these models on one data set from primate parietal and high-level visual cortices, and the other from rodent V1. The first dataset consisted of 14 simultaneously recorded neurons from the middle temporal visual (MT) and lateral intraparietal (LIP) area. These data were 50 1.4-second trials of a visual perceptual decision-making task \cite{yates2017functional}. In this task, random moving dots provided visual evidence towards left or right targets (choices). The trial contained a stimulus onset time, an evidence accumulation (decision making) period, and a decision. For the rodent data, spike times from 17 neurons were recorded during passive viewing 20 repeated 32-second trials of a gratings stimulus. The stimulus was a random flashing of gratings, with 8 orientations at fixed spatial and temporal frequencies. The gratings were presented for 4 seconds each.

\begin{figure}[t]%  figure placement: here, top, bottom, or page  
  \centering
  \includegraphics[width=1\textwidth]{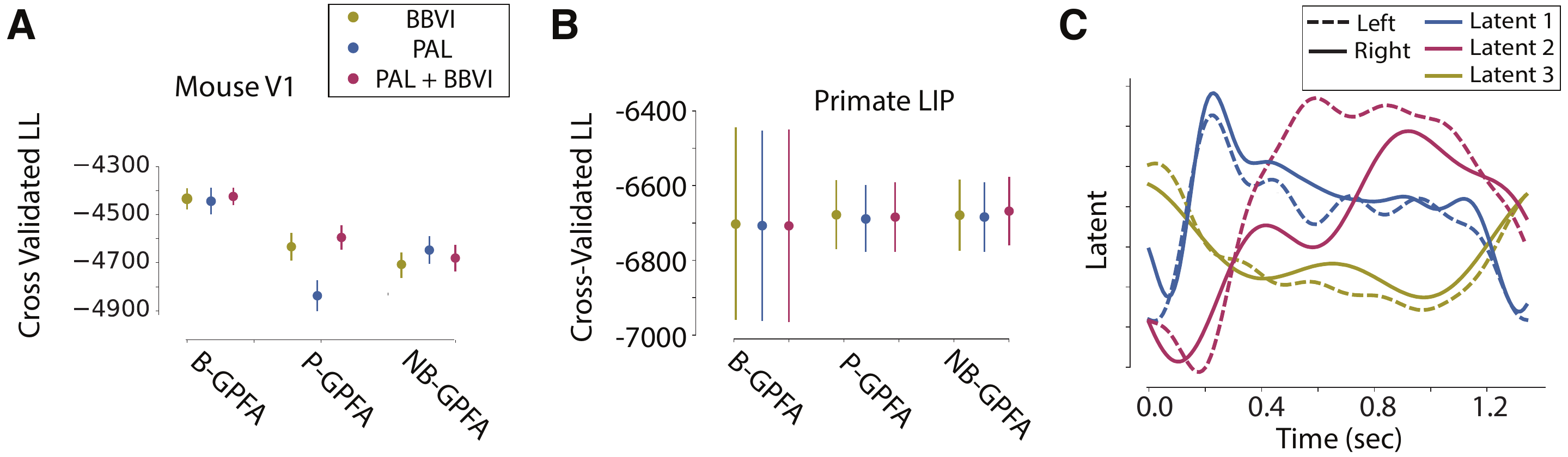} 
  \vspace{-2ex}
  \caption{Average cross validated log-likelihood on hold-out trials for PAL, BBVI and PAL + BBVI inference methods for count-GPFA models for mouse (\textbf{A}) and primate (\textbf{B}) data. (\textbf{C}) Latent structure underlying monkey LIP responses during left- and right-choice trials, showing that one latent dimension captured meaningful differences between the encoding of left and right choices.}
  \label{fig:OneRegNoise}
\end{figure}

 For the rodent and primate data, we asserted a latent dimensionality of three for all Count-GPFA models. Our PAL method demonstrated good empirical fits for real neural data (not shown) and good cross validation performance compared to BBVI  (Figure \ref{fig:OneRegNoise}A). However, for Poisson-GPFA, PAL performance was notably weaker. This was likely because these neurons were high-variance, exhibiting no activity for much of the trial, with some abrupt changes in firing throughout the trial. In this case, the exponential non-linearity is not accurately captured by the polynomial approximation. As noted in Figure 1, the exponential non-linearity in Poisson-GPFA is approximated least accurately of the three models, and operates over the smallest range. Thus, for this model, using PAL in conjunction with BBVI performed best. For the other two count models, Binomial and Negative Binomial GPFA, we found that our PAL estimation procedure is approximately as accurate as BBVI on these data. Interestingly, the count-GPFA model that had highest cross-validated log-likelihood was Binomial-GPFA, which is a count-model not often considered in neuroscience settings. %All count models performed better than standard GPFA on mean-squared error metrics for these data (not shown).
  %\vspace{-2ex}

  For the primate data, all GPFA models and inference methods exhibited equal performance (Figure \ref{fig:OneRegNoise}B), with a small bias favoring PAL-initialized negative binomial GPFA. The lack of major differences in performance here is likely because these data were high-spike rates neurons with many trials. This resulted in low-variance, stable rates that could be accurately captured by the PAL non-linearity and quickly and easily recovered using BBVI. %The differences in performance of count-GPFA predictions of neural rates are demonstrated in Figure \ref{fig:OneRegNoise}B. Here, the GPFA predictions are often negative, an impossibility with count-GPFA models. Additionally, GPFA overestimates smoothness of neural trajectories, attributing neural variability to observation noise rather than changes in rates. Count-GPFA models better capture situations where data are low-rate and involve abrupt changes in spiking.

%  For the primate data, latent dimensionality for these data was selected to be 3 dimensional for all models tested, verified again by cross validated log-likelihood. Mean squared error was calculated and reported as it was for the rodent data. 

To give insight into the scientific uses of this model, we show results of the Binomial-GPFA model fit to monkey LIP data. We fit a Binomial-GPFA model with 3 latent dimensions to two different subsets of the data: one consisting of the leftward choice trials and another consisting of rightward choice trials. We then compared the latents inferred for each condition in order to examine how the latent variables encode the animal's choice. 
%To do this, we aligned the three latent dimensions in each condition such that they span the same subspace. That is, we regressed one loadings matrix onto the other, and used that mapping to transform the latent space such that the latents could be meaningfully compared. 
Figure \ref{fig:OneRegNoise}C shows the latent structure of the neural data for each condition. Two of the latents are closely overlapping, which suggests the presence of shared structure across the two conditions. Interestingly, one latent (red) diverges near 400ms after trial onset, which falls into the portion of the trial where the animal is putatively making its choice. This suggests that this latent may encode the choice variable in these neural data, and is a promising future direction of further exploration for count-GPFA models.

\section{Conclusion}
We have a developed novel technique for learning Gaussian process factor analytic models with count observations using polynomial approximate log-likelihood (PAL) which allows for rapid closed-form evaluation of marginal likelihoods. We develop our PAL approach for three count-observation models: binomial, Poisson, and negative-binomial. In each case, our approximation can accurately estimate model parameters, and achieve good performance on both simulated and real neural data. PAL can additionally provide initial values for black box variational inference. Both PAL and BBVI have their own limitations -- PAL provides an approximation to the model non-linearity that is only accurate within a particular range, and BBVI inference procedure is sampling-based and can get stuck in local optima. Combining the procedures by using the PAL method to identify approximate hyperparameters can thus stabilize BBVI and make it more reliable, overcoming well-known BBVI optimization limitations. Overall, our PAL inference method is a novel approach to learning non-conjugate models that is fast and achieves high accuracy. %It is moreover able to extract low-dimensional latent structure on both simulated and real neural data. %We tested our various non-conjugate GPFA models on neural data and these count-GPFA models are comparable or better than traditional GPFA approaches, which do not often consider count noise. 

\section*{Acknowledgements}
SLK was supported by NIH grant F32MH115445-03, DMZ was supported by NIH grant T32MH065214 and JWP was supported by grants from the Simons Collaboration on the Global Brain (SCGB AWD543027), the NIH BRAIN initiative (NS104899 and R01EB026946), and a U19 NIH-NINDS BRAIN Initiative Award
(5U19NS104648). YY and SLS were supported by the NIH (R01EY024294 and R01NS091335), the NSF (1450824 and 1707287) the Simons Foundation (SCGB 325407) and the McKnight Foundation. The authors thank Adam Charles, Mikio Aoi, Nicholas Roy, and the anonymous reviewers for providing helpful comments.

\bibliographystyle{unsrt}
\bibliography{CountGPFA_arxiv.bib}

\begin{thebibliography}{10}

\bibitem{byron2009gaussian}
BM~Yu, JP~Cunningham, G~Santhanam, SI~Ryu, KV~Shenoy, and M~Sahani.
\newblock Gaussian-process factor analysis for low-dimensional single-trial
  analysis of neural population activity.
\newblock In {\em Adv neur inf proc sys}, pages 1881--1888, 2009.

\bibitem{cunningham2014dimensionality}
John~P Cunningham and B~M Yu.
\newblock Dimensionality reduction for large-scale neural recordings.
\newblock {\em Nature neuroscience}, 17(11):1500--1509, 2014.

\bibitem{lakshmanan2015extracting}
KC~Lakshmanan, PT~Sadtler, EC~Tyler-Kabara, AP~Batista, and BM~Yu.
\newblock Extracting low-dimensional latent structure from time series in the
  presence of delays.
\newblock {\em Neural computation}, 2015.

\bibitem{Archer15}
Evan Archer, Il~Memming Park, Lars Buesing, John Cunningham, and Liam Paninski.
\newblock Black box variational inference for state space models.
\newblock {\em stat}, 1050:23, 2015.

\bibitem{wu2017gaussian}
Anqi Wu, Nicholas~G Roy, Stephen Keeley, and Jonathan~W Pillow.
\newblock Gaussian process based nonlinear latent structure discovery in
  multivariate spike train data.
\newblock In I.~Guyon, U.~V. Luxburg, S.~Bengio, H.~Wallach, R.~Fergus,
  S.~Vishwanathan, and R.~Garnett, editors, {\em Advances in Neural Information
  Processing Systems 30}, pages 3499--3508. Curran Associates, Inc., 2017.

\bibitem{zhao2017variational}
Yuan Zhao and Il~Memming Park.
\newblock Variational latent gaussian process for recovering single-trial
  dynamics from population spike trains.
\newblock {\em Neural computation}, 29(5):1293--1316, 2017.

\bibitem{zhao2019stimulus}
Yuan Zhao, Jacob~Lachenmyer Yates, Aaron~Joseph Levi, Alexander~Christopher
  Huk, and Il~Memming Park.
\newblock Stimulus-choice (mis) alignment in primate mt cortex.
\newblock {\em bioRxiv}, 2019.

\bibitem{buesing2012spectral}
L~Buesing, J~H Macke, and M~Sahani.
\newblock Spectral learning of linear dynamics from generalised-linear
  observations with application to neural population data.
\newblock In {\em Adv neur inf proc sys}, pages 1682--1690, 2012.

\bibitem{macke2011empirical}
Jakob~H Macke, Lars Buesing, John~P Cunningham, M~Yu Byron, Krishna~V Shenoy,
  and Maneesh Sahani.
\newblock Empirical models of spiking in neural populations.
\newblock In {\em Advances in neural information processing systems}, pages
  1350--1358, 2011.

\bibitem{huggins2017pass}
Jonathan Huggins, Ryan~P Adams, and Tamara Broderick.
\newblock Pass-glm: polynomial approximate sufficient statistics for scalable
  bayesian glm inference.
\newblock In {\em Advances in Neural Information Processing Systems}, pages
  3614--3624, 2017.

\bibitem{Zoltowski18nips}
David~M Zoltowski and Jonathan~W Pillow.
\newblock Scaling the poisson glm to massive neural datasets through polynomial
  approximations.
\newblock In S.~Bengio, H.~Wallach, H.~Larochelle, K.~Grauman, N.~Cesa-Bianchi,
  and R.~Garnett, editors, {\em Advances in Neural Information Processing
  Systems 31}, pages 3521--3531. Curran Associates, Inc., 2018.

\bibitem{Charles18}
Adam~S Charles, Mijung Park, J~Patrick Weller, Gregory~D Horwitz, and
  Jonathan~W Pillow.
\newblock Dethroning the fano factor: A flexible, model-based approach to
  partitioning neural variability.
\newblock {\em Neural computation}, 30(4):1012--1045, 2018.

\bibitem{goris2014partitioning}
RLT Goris, JA~Movshon, and EP~Simoncelli.
\newblock Partitioning neuronal variability.
\newblock {\em Nature neuroscience}, 17(6):858--865, 2014.

\bibitem{linderman2016bayesian}
Scott Linderman, Ryan~P Adams, and Jonathan~W Pillow.
\newblock Bayesian latent structure discovery from multi-neuron recordings.
\newblock In {\em Advances in neural information processing systems}, pages
  2002--2010, 2016.

\bibitem{ranganath2014black}
Rajesh Ranganath, Sean Gerrish, and David Blei.
\newblock Black box variational inference.
\newblock In {\em Artificial Intelligence and Statistics}, pages 814--822,
  2014.

\bibitem{Gao15}
Yuanjun Gao, Lars Busing, Krishna~V Shenoy, and John~P Cunningham.
\newblock High-dimensional neural spike train analysis with generalized count
  linear dynamical systems.
\newblock In {\em Advances in neural information processing systems}, pages
  2044--2052, 2015.

\bibitem{Kara00}
P.~Kara, P.~Reinagel, and R.~C Reid.
\newblock Low response variability in simultaneously recorded retinal,
  thalamic, and cortical neurons.
\newblock {\em Neuron}, 27:636--646, 2000.

\bibitem{Maimon09}
Gaby Maimon and John~A Assad.
\newblock Beyond poisson: increased spike-time regularity across primate
  parietal cortex.
\newblock {\em Neuron}, 62(3):426--440, May 2009.

\bibitem{Pillow12}
Jonathan Pillow and James Scott.
\newblock Fully bayesian inference for neural models with negative-binomial
  spiking.
\newblock In P.~Bartlett, F.C.N. Pereira, C.J.C. Burges, L.~Bottou, and K.Q.
  Weinberger, editors, {\em Advances in Neural Information Processing Systems
  25}, pages 1907--1915, 2012.

\bibitem{duncker2018temporal}
Lea Duncker and Maneesh Sahani.
\newblock Temporal alignment and latent gaussian process factor inference in
  population spike trains.
\newblock {\em bioRxiv}, page 331751, 2018.

\bibitem{mason2002chebyshev}
John~C Mason and David~C Handscomb.
\newblock {\em Chebyshev polynomials}.
\newblock CRC Press, 2002.

\bibitem{blei2003latent}
David~M Blei, Andrew~Y Ng, and Michael~I Jordan.
\newblock Latent dirichlet allocation.
\newblock {\em Journal of machine Learning research}, 3(Jan):993--1022, 2003.

\bibitem{kingma2015variational}
Diederik~P Kingma, Tim Salimans, and Max Welling.
\newblock Variational dropout and the local reparameterization trick.
\newblock In {\em Advances in Neural Information Processing Systems}, pages
  2575--2583, 2015.

\bibitem{roeder2017sticking}
Geoffrey Roeder, Yuhuai Wu, and David Duvenaud.
\newblock Sticking the landing: An asymptotically zero-variance gradient
  estimator for variational inference.
\newblock {\em Advances in Neural Information Processing Systems}, 2017.

\bibitem{hoffman2013stochastic}
Matthew~D Hoffman, David~M Blei, Chong Wang, and John Paisley.
\newblock Stochastic variational inference.
\newblock {\em The Journal of Machine Learning Research}, 14(1):1303--1347,
  2013.

\bibitem{salimbeni2018natural}
Hugh Salimbeni, Stefanos Eleftheriadis, and James Hensman.
\newblock Natural gradients in practice: Non-conjugate variational inference in
  gaussian process models.
\newblock {\em arXiv preprint arXiv:1803.09151}, 2018.

\bibitem{yates2017functional}
Jacob~L Yates, Il~Memming Park, Leor~N Katz, Jonathan~W Pillow, and Alexander~C
  Huk.
\newblock Functional dissection of signal and noise in mt and lip during
  decision-making.
\newblock {\em Nature neuroscience}, 20(9):1285, 2017.

\end{thebibliography}

% \end{figure*}

\end{document}